\documentclass{article}
\usepackage{amsmath,amsthm,amssymb}
\usepackage{bbm}
\usepackage{mathtools}
\usepackage{graphicx}
\usepackage{float}
\usepackage{xspace}

\usepackage[dvipsnames]{xcolor}
\usepackage[toc, page]{appendix}
\usepackage[framemethod=TikZ]{mdframed}

\usepackage{natbib}

\usepackage{nicefrac}       %
\usepackage{microtype}      %
\usepackage[acronym, nopostdot, nogroupskip, nohypertypes={acronym,notation}]{glossaries}
\makeglossaries
\glsaddall
\usepackage{cancel}
\usepackage{dirtytalk}
\usepackage{wrapfig}
\usepackage{dsfont} %
\usepackage{placeins} %
\usepackage{rotating} %

\usepackage{hyperref}
\hypersetup{
    colorlinks=true,
    linkcolor=[rgb]{0.709, 0.070, 0.105},
    citecolor=[rgb]{0.709, 0.070, 0.105},
    filecolor=magenta,      
    urlcolor=[rgb]{0.709, 0.070, 0.105},
}
\usepackage[nameinlink]{cleveref} %

\usepackage{thmtools}
\declaretheoremstyle[
notefont=\bfseries, 
notebraces={(}{)},
bodyfont=\normalfont\itshape,
headformat=\NAME~\NUMBER:\NOTE,
spacebelow=\parsep,
spaceabove=\parsep,
mdframed={
    backgroundcolor=blue!2, 
    linecolor=blue!20, 
    innertopmargin=6pt,
    roundcorner=20pt, 
    innerbottommargin=6pt, 
    skipabove=\parsep, 
    skipbelow=\parsep,
    nobreak=true} 
]{BoldParanthesesNoted}

\usepackage{algorithm}

\usepackage{algorithmic}
\let\Algorithm\algorithm
\renewcommand\algorithm[1][]{\Algorithm[#1]\setstretch{1.2}}

\usepackage{listings}
\usepackage{xparse}
\NewDocumentCommand{\codeword}{v}{%
\texttt{\textcolor{OliveGreen}{#1}}%
}

\usepackage{tikz,tikz-3dplot}
\tdplotsetmaincoords{60}{120}
\usetikzlibrary{topaths,calc,bayesnet, arrows}
\usepackage{subcaption}

\usepackage{caption} 
\usepackage{booktabs}

\usepackage{multirow}
\usepackage{multicol}
\usepackage{makecell} %

\newacronym{BO}{BO}{Bayesian optimisation}
\newacronym{MT}{MT}{multi-task}
\newacronym{MO}{MO}{multi-objective}
\newacronym{MF}{MF}{multi-fidelity}
\newacronym{MESMO}{MESMO}{max-value entropy search for multi-objective optimisation}
\newacronym{EI}{EI}{expected information}
\newacronym{ES}{ES}{entropy search}
\newacronym{PES}{PES}{predictive entropy search}
\newacronym{MVES}{MVES}{max-value entropy search}
\newacronym{MOBO}{MOBO}{multi-objective Bayesian optimisation}
\newacronym{PHV}{PHV}{Pareto hypervolume}
\newacronym{RKHS}{RKHS}{Reproducing Kernel Hilbert Space}
\newacronym{IPM}{IPM}{integral probability metric}
\newacronym{MCMC}{MCMC}{Markov chain Monte-Carlo}
\newacronym{HMC}{HMC}{Hamiltonian Monte-Carlo}
\newacronym{SVI}{SVI}{stochastic variational inference}
\newacronym{SVGD}{SVGD}{Stein variational gradient descent}
\newacronym{VI}{VI}{variational inference}
\newacronym{KLD}{KLD}{Kullback-Leibler divergence}
\newacronym{IID}{IID}{independent and identically distributed}
\newacronym{ELBO}{ELBO}{evidence lower bound}
\newacronym{RBF}{RBF}{radial basis function}
\newacronym{ARD}{ARD}{Automatic Relevance Determination }
\newacronym{KSD}{KSD}{Kernel Stein Discrepancy}
\newacronym{PD}{PD}{positive definite}
\newacronym{MMD}{MMD}{Maximum mean discrepancy}
\newacronym{MAP}{MAP}{maximum a posteriori}
\newacronym{NOx}{NOx}{nitrogen oxide}
\newacronym{VFF}{VFF}{variational Fourier feature}
\newacronym{SPSD}{SPSD}{symmetric positive-semidefinite}
\newacronym{SPDE}{SPDE}{stochastic partial differential equation}
\newacronym{INLA}{INLA}{integrated nested Laplace approximation}
\newacronym{BNN}{BNN}{Bayesian neural network}
\newacronym{MCTS}{MCTS}{Monte-Carlo tree search}
\newacronym{UCB}{UCB}{upper confidence bound}
\newacronym{GPLVM}{GPLVM}{Gaussian process latent variable model}
\newacronym{FP}{FP}{false positive}
\newacronym{TP}{TP}{true positive}
\newacronym{FN}{FN}{false negative}
\newacronym{TN}{TN}{true negative}
\newacronym{ECE}{ECE}{expected calibration error}

\newacronym{tvd}{TVD}{total variation distance}
\newglossaryentry{GP}
{
  name={GP},
  description={Gaussian process},
  first={\glsentrydesc{GP} (\glsentrytext{GP})},
  plural={GPs},
  descriptionplural={Gaussian processes},
  firstplural={\glsentrydescplural{GP} (\glsentryplural{GP})}
} 

\newglossaryentry{GNN}
{
  name={GNN},
  description={graph neural network},
  first={\glsentrydesc{GNN} (\glsentrytext{GNN})},
  plural={GNNs},
  descriptionplural={graph neural networks},
  firstplural={\glsentrydescplural{GNN} (\glsentryplural{GNN})}
} 

\newglossaryentry{DGP}
{
  name={DGP},
  description={deep Gaussian process},
  first={\glsentrydesc{DGP} (\glsentrytext{DGP})},
  plural={DGPs},
  descriptionplural={deep Gaussian processes},
  firstplural={\glsentrydescplural{DGP} (\glsentryplural{DGP})}
}

\newglossaryentry{RFF}
{
  name={RFF},
  description={Random Fourier Feature},
  first={\glsentrydesc{RFF} (\glsentrytext{RFF})},
  plural={RFFs},
  descriptionplural={Random Fourier Features},
  firstplural={\glsentrydescplural{RFF} (\glsentryplural{RFF})}
} 

\newglossaryentry{GMRF}
{
  name={GMRF},
  description={Gaussian Markov random field},
  first={\glsentrydesc{GMRF} (\glsentrytext{GMRF})},
  plural={GMRFs},
  descriptionplural={Gaussian Markov random fields},
  firstplural={\glsentrydescplural{GMRF} (\glsentryplural{GMRF})}
} 

\newglossaryentry{RV}
{
  name={RV},
  description={random variable},
  first={\glsentrydesc{RV} (\glsentrytext{RV})},
  plural={RVs},
  descriptionplural={random variables},
  firstplural={\glsentrydescplural{RV} (\glsentryplural{RV})}
}

\newglossaryentry{KDPP}
{
  name={KDPP},
  description={$k$-determinantal point process},
  first={\glsentrydesc{KDPP} (\glsentrytext{KDPP})},
  plural={KDPPs},
  descriptionplural={$k$-determinantal point processes},
  firstplural={\glsentrydescplural{KDPP} (\glsentryplural{KDPP})}
} 

\newglossaryentry{MOGP}
{
  name={MOGP},
  description={multi-output Gaussian process},
  first={\glsentrydesc{MOGP} (\glsentrytext{MOGP})},
  plural={MOGPs},
  descriptionplural={multi-output Gaussian processes},
  firstplural={\glsentrydescplural{MOGP} (\glsentryplural{MOGP})}
} 

\newacronym{DTC}{DTC}{Deterministic Training Conditional}
\newacronym{FITC}{FITC}{Fully Independent Training Conditional}
\newacronym{PITC}{PITC}{Partially Independent Training Conditional}
\newacronym{VFE}{VFE}{Variational Free Energy}
\newacronym{KPMF}{KPMF}{kernel probabilistic matrix factorisation}
\newacronym{PMF}{PMF}{probabilistic matrix factorisation}
\newacronym{RMSE}{RMSE}{root-mean-square error}

\usepackage{xcolor}

\definecolor{dkgreen}{rgb}{0,0.6,0}
\definecolor{gray}{rgb}{0.5,0.5,0.5}
\definecolor{mauve}{rgb}{0.58,0,0.82}

\definecolor{midnightblue}{rgb}{0.15, 0.15, 0.5}

\newcommand{\tabref}[1]{Table \ref{#1}}

\newcommand{\secref}[1]{Section \ref{#1}}

\newcommand{\x}{\mathbf{x}}
\newcommand{\z}{\mathbf{z}}

\newcommand{\bu}{\mathbf{u}}

\newcommand{\y}{y}
\newcommand{\by}{\mathbf{y}}

\DeclareMathOperator*{\argmax}{arg\,max}
\DeclareMathOperator*{\argmin}{arg\,min}

\newcommand{\GP}{\mathcal{GP}}

\newcommand{\bfx}{\mathbf{f}_{\star}}

\newcommand{\Kxx}{\mathbf{K}_{\mathbf{\x\x}}}
\newcommand{\KVV}{\mathbf{K}_{VV}}

\newcommand{\Kzz}{\mathbf{K}_{\mathbf{\mathbf{zz}}}}

\newcommand{\KL}{\text{KL}}

\newcommand{\f}{\mathbf{f}}
\newcommand{\given}{\,|\,}
\newcommand{\btheta}{\boldsymbol\theta}

\newcommand{\cD}{\mathcal{D}}

\newcommand{\cG}{\mathcal{G}}
\newcommand{\cH}{\mathcal{H}}

\newcommand{\cN}{\mathcal{N}}
\newcommand{\cO}{\mathcal{O}}

\newcommand{\cX}{\mathcal{X}}

\newcommand{\cW}{\mathcal{W}}

\newcommand{\bm}{\mathbf{m}}

\newcommand{\bbR}{\mathbb{R}}

\renewcommand{\KL}{\operatorname{KL}}

\newcommand{\TruePos}{\operatorname{TP}}
\newcommand{\FalsePos}{\operatorname{FP}}

\newcommand{\FalseNeg}{\operatorname{FN}}

\title{\textbf{Gaussian Processes on Hypergraphs}}
\author{Thomas Pinder$^{1}$, Kathryn Turnbull$^{1}$ \\ 
Christopher Nemeth$^1$, David Leslie$^1$ \\ \vspace{-0.2cm}
{\footnotesize $^1$Department of Mathematics and Statistics, Lancaster University, UK} \\ \vspace{-0.2cm}
}
\date{\today}

\begin{document}

\maketitle

\begin{abstract}
We derive a Mat\'ern Gaussian process (GP) on the vertices of a hypergraph. This enables estimation of regression models of observed or latent values associated with the vertices, in which the correlation and uncertainty estimates are informed by the hypergraph structure. We further present a framework for embedding the vertices of a hypergraph into a latent space using the hypergraph GP. Finally, we provide a scheme for identifying a small number of representative inducing vertices that enables scalable inference through sparse GPs. We demonstrate the utility of our framework on three challenging real-world problems that concern multi-class classification for the political party affiliation of legislators on the basis of voting behaviour, probabilistic matrix factorisation of movie reviews, and embedding a hypergraph of animals into a low-dimensional latent space.
\end{abstract}

\section{Introduction} \label{sec:intro}
\glspl{GP} are a popular type of stochastic process commonly used in machine learning for modeling distributions over real-valued functions \citep{rasmussen_gaussian_2006}. In recent years, \glspl{GP} have been successfully used in areas such as optimisation \citep{mockus_bayesian_2012}, variance-reduction \citep{oates_control_2017} and robotics \citep{deisenroth_gaussian_2015}. This recent blossoming is due to the GP's ability to model complex functional relationships with accompanying well-calibrated uncertainty estimates.

Careful consideration of the underlying data's representation has facilitated more expressive \gls{GP} priors to be constructed. Examples of this include the modelling of regular %
time-series models \citep{sarkka_applied_2019, adam_doubly2020}, spatial data observed on a uniform grid \citep{solin_hilbert_2019} and multivariate outputs where inter-variable correlations exists \citep{alvarez_kernels_2012}. In this work, we develop and implement a GP prior for situations where relationships between the explanatory variables in the data admit a \textit{hypergraph} structure.%

As a motivating example, consider modelling the party affiliation of a group of politicians using legislative information. It is natural to expect that politicians who collaborate on a piece of legislation share some similarity. Since any set of politicians may work together, this dependence can be represented by a hypergraph where each politician is a vertex and each hyperedge corresponds to a piece of legislation. Whilst we may also describe this data structure as a graph of pairwise collaborations, this representation cannot express the full structural information encoded in the hypergraph where collaborations occur between two, or more, politicians. When the relationship between observations is represented by a hypergraph, the Euclidean measure of distance between pairs of observations is no longer appropriate and we must instead build a distance function using the structure of the underlying hypergraph. %

The contributions of this paper can be seen as a generalisation of \cite{borovitskiy_matern_2020}  whereby we 
introduce a hypergraph kernel to 
allow higher-order interactions to be represented %
in the \gls{GP} prior. This results in a prior distribution that is more representative of the underlying data's structure, and we find that it leads to significant improvements in the \gls{GP} posterior's inferential performance. In addition to this, we also present a novel approach for embedding hypergraphs in a lower dimensional latent space through a \gls{GPLVM}, present a new scheme for selecting inducing points in a hypergraph \gls{GP} and finally show how our hypergraph \gls{GP} can be used within a \gls{PMF} framework.

\section{Background}\label{sec:back}

\subsection{Gaussian processes}\label{sec:back:gp}
For a set $\cX$, a stochastic process $f: \cX \rightarrow \mathbb{R}$ is a \gls{GP} if for any finite collection $X = \{\x_1, \x_2, \ldots , \x_n\} \subset \cX$ the random variable $\f :=f(X)$ is distributed jointly Gaussian \citep{rasmussen_gaussian_2006}. We write a \gls{GP} $f \sim \GP(\mu, k)$ with mean function $\mu: \cX \rightarrow \mathbb{R}$ and $\btheta$-parameterised kernel $k: \cX \times \cX \rightarrow \mathbb{R}$ where $k$ gives rise to the Gram matrix $\Kxx$ such that the $(i, j)^{\text{th}}$ entry is computed by $[\Kxx]_{i, j} = k(\x_i, \x_j)$. In standard expositions of stationary Gaussian processes, with $\cX=\bbR^d$, $k(\x_i,\x_j)$ is a function of the distance between $\x_i$ and $\x_j$, whereas this article introduces a kernel that is evaluated on vertices of a hypergraph. Without loss of generality, we assume $\mu(\x) := 0$ for all $\x$ and drop dependency on $\btheta$ for notational convenience.

Assume an observed dataset $\cD = \{\x_i,y_i\}_{i=1}^N$ of $N$ training example pairs with inputs $\x_i \in \cX$ and outputs $y_i\in\bbR$ distributed according to $\y\given \f \sim \prod_{i=1}^N p(y_i\given f_i)$. Then the predictive density for latent function values $\bfx$ at a finite collection of $N^{\star}$ test points $X^{\star}=\{\x_1^{\star}, \ldots, \x_{N^{\star}}^{\star}\}$ is given by 
\begin{equation}
    \label{equn:GPPosterior}
    p(\bfx\given\cD)  = \int p\left(\bfx ,\f \given \cD\right)\,\mathrm{d} \mathbf{f}  =\int p\left(\bfx | \f \right) p(\f | \cD) \,\mathrm{d}  \mathbf{f},
\end{equation}
where $p(\f |\cD) \propto p(\by|\f)p(\f)$ is the posterior density of $\f$ given the training data. When the observations are distributed according to a Gaussian likelihood, both $p(\bfx\given\f)$ and $p(\f\given\cD)$ are Gaussian densities. The marginal distribution of a Gaussian is also Gaussian and so the posterior in \Cref{equn:GPPosterior} is a Gaussian distribution with a closed-form expression for the mean and covariance. When the likelihood function is non-Gaussian, the posterior distribution over the latent function $p(\f\given\cD)$ has no closed-form expression and inference for \Cref{equn:GPPosterior} is no longer tractable. Popular solutions that address this problem involve approximating $p(\f\given\cD)$ through a Laplace approximation \citep{williams_bayesian_1998} or variational inference \citep{opper_variational_2008}. Alternatively, the latent function's values can be inferred directly using either \gls{MCMC} \citep{murray_elliptical_2010} or Stein variational gradient descent \citep{pinder_stein_2020}.

The primary cost incurred when computing the \gls{GP} posterior is the inversion of the Gram matrix $\Kxx$. Sparse approximations \citep{snelson2005sparse} temper this cubic in $N$ cost through the introduction of a set of $J$ \textit{inducing points} $Z = \{\z_1, \z_2, \ldots, \z_J\} \in \cX$ and corresponding function outputs $\bu = f(Z)$. Letting $J \ll N$, this approach reduces the computational requirement to $\cO(NJ^2)$ from $\cO(N^3)$. Following \cite{titsias2009}, we augment the posterior distribution with $\bu$ to give $p(\f, \bu \given \by) = p(\f\given\bu, \by)p(\bu\given\by)$ and then introduce the variational approximation $q(\f, \bu) = p(\f \given \bu) q(\bu)$. Constraining $q$ to be a multivariate Gaussian $q(\bu) = \cN(\bm, S)$ and collecting our parameters of interest $\psi = \{Z, \bm, S, \btheta\}$, we can determine the \textit{optimal} parameters $\psi^{\star}$ for the variational approximation as 
\begin{align}
    \label{equn:VFEObjective}
    \psi^{\star} & = \argmin_{\psi \in \Psi} \KL(q(\f, \bu) || p(\f, \bu \given \by)) \nonumber\\
    & \geq \argmax_{\psi \in \Psi} \sum_{n=1}^N \mathbb{E}_{q(\bu, \f)}\left[\log p(y_i \given f_i)\right] - \KL(q(\bu) || p(\bu))
\end{align}
where $\Psi$ is the space of all possible parameters and $p(\bu) = \cN(\bu\given \mathbf{0},\Kzz)$ such that $[\Kzz]_{i, j} = k(\z_i, \z_j)$. The quantity in \Cref{equn:VFEObjective} is commonly referred to as the \gls{ELBO} and is equal to the marginal log-likelihood when $ \KL(q(\bu) || p(\bu))=0$. For a full introduction to variational sparse \glspl{GP} see \cite{leibfried_2020_sparse}.

\subsection{Hypergraphs}\label{sec:model:hypergraphs}
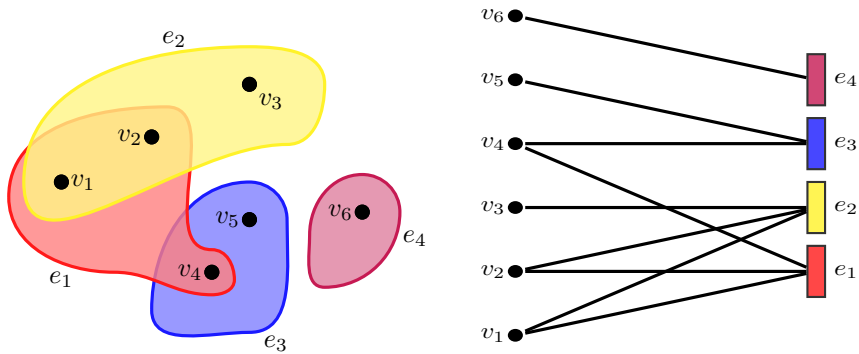
\begin{figure}[ht]
\centering
\begin{subfigure}{0.5\textwidth}
\begin{tikzpicture}
    \node (v1) at (0,1.7) {};
    \node (v2) at (2.5,3) {};
    \node (v3) at (2,0.5) {};
    \node (v4) at (2.5,1.2) {};
    \node (v5) at (4,1.3) {};
    \node (v6) at (1.2,2.3) {};

    \begin{scope}[fill opacity=0.8]
    \filldraw[ color=blue!90, fill=blue!50, very thick] ($(v3)+(-0.8,-0.5)$) %
        to[out=90,in=180] ($(v4) + (0,0.5)$) %
        to[out=0,in=90] ($(v4) + (0.5,0)$) %
        to[out=270,in=0] ($(v4) + (0,-1.5)$)
        to[out=180,in=270] ($(v3)+(-0.8,-0.5)$);
    \filldraw[color=red!90, fill=red!50, very thick] ($(v1)+(-0.7,-0.2)$) 
        to[out=90,in=180] ($(v6) + (0.2,0.4)$) 
        to[out=0,in=180] ($(v3) + (0,0.3)$)
        to[out=0,in=90] ($(v3) + (0.3,-0.1)$)
        to[out=270,in=0] ($(v3) + (0,-0.3)$)
        to[out=180,in=0] ($(v3) + (-1.3,0)$)
        to[out=180,in=270] ($(v1)+(-0.7,-0.2)$);
    \filldraw[ color=yellow!90, fill=yellow!60, very thick] ($(v1)+(-0.5,0)$) %
        to[out=90,in=180] ($(v2) + (0,0.5)$) %
        to[out=0,in=90] ($(v2) + (1,0)$) %
        to[out=270,in=0] ($(v2) + (0.5,-0.8)$)
        to[out=180,in=270] ($(v1)+(-0.5,0)$);

    \filldraw[ color=purple!90, fill=purple!50, very thick] ($(v5)+(-0.7,-0.3)$) %
        to[out=90,in=180] ($(v5) + (0,0.5)$) %
        to[out=0,in=90] ($(v5) + (0.5,0)$) %
        to[out=270,in=0] ($(v5) + (-0.5,-1.0)$)
        to[out=180,in=270] ($(v5)+(-0.7,-0.3)$);
    \end{scope}

    \foreach \v in {1,2,...,6} {
        \fill (v\v) circle (0.1);
    }

    \fill (v1) circle (0.1) node [right] {$v_1$};
    \fill (v2) circle (0.1) node [below right] {$v_3$};
    \fill (v3) circle (0.1) node [left] {$v_4$};
    \fill (v4) circle (0.1) node [left] {$v_5$};
    \fill (v5) circle (0.1) node [left] {$v_6$};
    \fill (v6) circle (0.1) node [left] {$v_2$};

    \node at (0.0,0.4) {$e_1$};
    \node at (1.5,3.6) {$e_2$};
    \node at (2.85,-0.45) {$e_3$};
    \node at (4.7,0.95) {$e_4$};
\end{tikzpicture}
\end{subfigure} 
\begin{subfigure}{0.4\textwidth}
\begin{tikzpicture}[thick,yscale=0.85, every node/.style={transform shape}]
        \node (v1) at (0,0) {};
        \node (v2) at (0, 1) {};
        \node (v3) at (0,2) {};
        \node (v4) at (0,3) {};
        \node (v5) at (0,4) {};
        \node (v6) at (0,5) {};
        \node (v7) at (4,1) [draw,thick,minimum width=0.2cm,minimum height=0.8cm, fill=red!90, opacity=0.8] {};
        \node (v8) at (4,2) [draw,thick,minimum width=0.2cm,minimum height=0.8cm, fill=yellow!90, opacity=0.8] {};
        \node (v9) at (4,3) [draw,thick,minimum width=0.2cm,minimum height=0.8cm, fill=blue!90, opacity=0.8] {};
        \node (rect) (v10) at (4,4) [draw,thick,minimum width=0.2cm,minimum height=0.8cm, fill=purple!90, opacity=0.8] {};

        \path [-, very thick] (v1) edge node[left] {} (v7);
        \path [-, very thick] (v1) edge node[left] {} (v8);
        \path [-, very thick] (v2) edge node[left] {} (v7);
        \path [-, very thick] (v2) edge node[left] {} (v8);
        \path [-, very thick] (v3) edge node[left] {} (v8);
        \path [-, very thick] (v4) edge node[left] {} (v7);
        \path [-, very thick] (v4) edge node[left] {} (v9);
        \path [-, very thick] (v5) edge node[left] {} (v9);
        \path [-, very thick] (v6) edge node[left] {} (v10);
        
        \fill (v1) circle (0.1) node [left] {$v_1$};
        \fill (v2) circle (0.1) node [left] {$v_2$};
        \fill (v3) circle (0.1) node [left] {$v_3$};
        \fill (v4) circle (0.1) node [left] {$v_4$};
        \fill (v5) circle (0.1) node [left] {$v_5$};
        \fill (v6) circle (0.1) node [left] {$v_6$};
        
        \node  at (4.4, 1) {$e_1$};
        \node at (4.4, 2) {$e_2$};
        \node at (4.4, 3) {$e_3$};
        \node at (4.4, 4) {$e_4$};
\end{tikzpicture}
\end{subfigure}
\caption{A hypergraph comprised of four hyperedges among six vertices and the corresponding bipartite representation. See Section 2.1.2 in \cite{bretto2013} for details of this relationship.}
\label{fig:hypergraphSchematic}
\end{figure}

A \textit{hypergraph} \citep{bretto2013} is a generalisation of a graph in which interactions  occur among an arbitrary set of nodes (see \Cref{fig:hypergraphSchematic}). This type of data provides a flexible and descriptive framework to encode higher-order relationships within a graph-like structure.
Formally, we let $\cG = \{V, E\}$ be a hypergraph with vertices $V = \{v_1, \ldots , v_N\}$ and \textit{hyperedges} $E = \{e_1, \ldots, e_M\}$. Each hyperedge $e_i \in E$ corresponds to a subset of 
vertices from $V$ with no repeated elements. A hyperedge $e_i$ is said to be \textit{weighted} when it has a positive value $w(e
_i)$ assigned to it and \textit{incident} to a vertex $v_j$ if $v_j \in e_i$. We represent a hypergraph via a $\lvert N \rvert \times \lvert M \rvert$ incidence matrix $H$ with $(j,i)^{\text{th}}$ entry equal to 1 if $e_i$ is incident to $v_j$ and $0$ otherwise. 

Every hypergraph also admits a bipartite representation in which each hyperedge is assigned a node, and an edge from a node vertex to a hyperedge vertex indicates incidence (see Figure \ref{fig:hypergraphSchematic}). Since a hypergraph generalises the graph representation, it follows that these structures coincide when each hyperedge contains precisely two nodes. Whilst we may obtain a graph from a given hypergraph (see Appendix \ref{app:exp:graphRepr}), this inevitably loses structural information which cannot be recovered.

The Laplacian of a hypergraph is an important matrix representation which can be used to study hypergraph properties. This matrix has proven useful as part of machine learning algorithms, such as spectral clustering, and, due to its importance, several authors have proposed analogous hypergraph Laplacians (see \cite{agarwal2006higher}, \cite{battiston_networks_2020} for examples). We rely on the hypergraph Laplacian proposed by \cite{zhou_learning_2006} for spectral clustering, which is given by 
\begin{equation}
    \label{equn:HypergraphLaplacian}
    \Delta = I - D_v^{\nicefrac{-1}{2}}HD_e^{-1}H^{\top}D_v^{\nicefrac{-1}{2}},
\end{equation}
where $D_v$ and $D_e$ are diagonal matrices with non-zero entries containing the node and hyperedge degrees, respectively. We write
\begin{align}
    \label{equn:hypergraphDegreeMatrices}
    [D_v]_{i, i} = \sum_{e \in E}w(e)h(v_i, e) \qquad [D_e]_{i,i} = \sum_{v\in V}h(v, e_i),
\end{align}
where $h(v, e)=1$ if $v$ is incident to $e$ i.e., $v\in e$. In this work we only considered unweighted hypergraphs (i.e.\ $w(e_i)=1$ for $1\leq i\leq M$) however extending our framework to accommodate weighted hyperedges poses no challenge.

\section{Gaussian process on hypergraphs}\label{sec:model}
We wish to define a \gls{GP} prior for functions over the space containing the vertices of a hypergraph i.e., $\cX=V$. We do so by building from a \gls{SPDE} formulation of a \gls{GP} prior over Euclidean spaces.

\paragraph{\gls{SPDE} kernel representations}Defining kernel functions in terms of an \gls{SPDE} provides a flexible framework that facilitates the derivation of kernels on non-Euclidean spaces. To see this, we consider the \gls{SPDE}
\begin{align}
    \label{equn:WhiteNoiseSolution}
    \mathcal{T}f = \cW %
\end{align}
where $\mathcal{T}:\cX\rightarrow\cH$ is a bounded linear operator on a Hilbert space $\cH$, $f$ is a zero-mean Gaussian random field and $\cW$ a white noise process \citep{lototsky2017stochastic}. For a Euclidean domain i.e., $\cX=\bbR^d$, \cite{whittle_stochastic} has shown that the Mat\'ern kernel satisfies \Cref{equn:WhiteNoiseSolution} through
\begin{align}
    \label{equn:MaternSPDE}
    \left(\frac{2\nu}{\ell^2} + L \right)^{\nicefrac{\nu}{2} + \nicefrac{d}{4}}f= \mathbf{\cW},
\end{align}
where $L$ is the Laplace operator $Lf(\x) =\nabla \cdot \nabla f(\x)$; the divergence of the gradient of $f$, and $\nu\in\bbR_{>0}$ and $\ell\in\bbR_{>0}$ are the smoothness and lengthscale parameters of the Mat\'ern kernel respectively.

\paragraph{Hypergraph domain}
We transfer the construction \Cref{equn:MaternSPDE} into the hypergraph domain by taking $\cW$ to be a multivariate spherical Gaussian and replacing the Laplace operator $L$ with the hypergraph Laplacian $\Delta$. (We can also drop the term $d/4$ in the exponent, since it is not required for smoothness concerns in the discrete space $\cX$.) Since we have a discrete space $\cX$, \Cref{equn:MaternSPDE} becomes 
\begin{align}
    \label{equn:HypergaphKernelSPDE}
     \left(\frac{2\nu}{\ell^2} + \Delta \right)^{\nicefrac{\nu}{2}}
     \f = \cW,
\end{align}
where %
$\f$ is the vector of function values at the vertices of the hypergraph. We can rearrange \Cref{equn:HypergaphKernelSPDE} to give the hypergraph \gls{GP} prior 
\begin{align}
    \label{equn:GraphGPPrior}
    \f \sim \cN\left(0, \left(\frac{2\nu}{\ell^2} + \Delta \right)^{-\nu}\right).
\end{align}

\paragraph{Hypergraph dual} For a given hypergraph, we can construct a hypergraph dual whose vertices correspond to the hyperedges in the original hypergraph. This relationship is made clear through the bipartite representation in \Cref{fig:hypergraphSchematic}. We can leverage this duality to facilitate inference on either the vertices or hyperedges in our initial hypergraph, and we make use of this feature in \Cref{sec:exp:kpmf}.

\paragraph{Connection to graph \glspl{GP}}If we consider hypergraphs of order 2, i.e. $| e | = 2$ for all $e \in E$, then we recover the \gls{GP} formulated in \cite{borovitskiy_matern_2020} as a special case of the hypergraph GP. 

\subsection{Sparse hypergraph GPs}\label{sec:model:inducingPoints}
Revisiting the \gls{ELBO} term in \Cref{equn:VFEObjective}, we can see that for \glspl{GP} defined on continuous domains, the inducing points $Z=\{\mathbf{z}_1, \mathbf{z}_2, \ldots, \mathbf{z}_J\}$ are treated as a model parameter that we optimise. Despite this, poor initialisation can lead to significantly slower convergence and often a poor variational approximation \citep{burt_convergence_2020}. For hypergraphs, such an optimisation is not possible as the notion of an inducing point is now a discrete quantity that corresponds to a specific vertex's index in the hypergraph. We hereafter refer to inducing points in a hypergraph as \textit{inducing vertices}. As we cannot optimise our inducing vertices, it is important that we initialise them effectively as a poor initialisation cannot be rectified through an optimisation scheme. 

Our approach is a three-stage process where we first assign each vertex a globally measured \textit{importance} score $\gamma \in[0, 1]$. We use this measure to ensure that the most influential vertices in our hypergraph are present in the inducing set. Second, we cluster the vertices of the hypergraphs by using $k$-means clustering on the Laplacian matrix's eigenvalues. Finally, using the clustered vertices, we first sample a cluster with probability equal to the the cluster's normalised size, then select the vertex with the largest importance score in the respective cluster. 

\paragraph{Vertex importance}To compute importance scores for each vertex we project our hypergraph onto a graph and calculate the eigencentrality \citep[see][]{kolaczyk2009} of each vertex. We first calculate the adjacency matrix $A_v = HH^{\top}$ with diagonal entries equal to the vertex degrees and off-diagonals equal to the number of times two vertices are incident to one another. If we now define the stochastic matrix $Q = A_v D_v^{-1}$ that scales the adjacency matrix by the vertices' degree, then we can obtain centrality measures through $\lambda\boldsymbol{\gamma}=Q\boldsymbol{\gamma}$. We can determine $\boldsymbol{\gamma}$ by identifying the eigenvector that corresponds to the largest eigenvalue $\lambda$, and the centrality score for the $v^{\text{th}}$ vertex is then given by $\gamma_v$. We note that $H$ and $D_v$ are non-negative matrices and therefore $Q$ is also non-negative. Consequently, the Perron-Frobenius theorem guarantees the existence of a unique $\boldsymbol{\gamma}$.

\paragraph{Clustering} To cluster the vertices of our hypergraph, $k$-means clustering is carried out on the eigenvectors of the hypergraph's Laplacian matrix \Cref{equn:HypergraphLaplacian}. In the absence of informative prior information, such as the number of observation classes, we suggest selecting a conservative value of $k\ll J$.

\paragraph{Inducing vertex selection}
A set of $J$ inducing vertices can now be identified by a two-step procedure. Let $k_i$ denote the $i^{\text{th}}$ cluster that contains $\lvert k_i \rvert$ vertices and $\operatorname{cat}$ be the categorical distribution with $k$ categories with corresponding probabilities $\mathbf{p}=\{p_1, p_2, \ldots, p_k\}$. The $j^{\text{th}}$ element of $Z$ can then be selected by
\begin{align}
    \label{equn:InducingVertexSelection}
    \text{Sample:}&\  s\sim \operatorname{cat}(\mathbf{p}), \ \text{where} \ \mathbf{p}\propto\left\lbrace\frac{\lvert k_1 \rvert}{N}, \frac{\lvert k_2 \rvert}{N},\ldots,  \frac{\lvert k_s \rvert}{N}\right\rbrace \nonumber\\
    \text{Select:}&\ z_j = \argmax_{v \in s\given v \notin Z_{1:(j-1)}}\gamma_v.
\end{align}
This process is repeated until $\lvert Z\rvert = J$.

Unlike uniform random sampling, the approach outlined here is more robust to situations where a class imbalance in the observations' labels is present due to the clustering step performed. Further, we are able to identify the most influential vertices within the graph through the assignment of an eigen-based importance measure to each vertex; a metric which captures a vertex's \textit{global} importance. Despite this, the framework presented here is highly modular, and the practitioner is free to use alternative clustering methods, such as the popular DBSCAN \citep{Ester_1996_DBSCAN}, or alternative centrality measures on the hypergraph's vertices \citep{benson_2019_centrality}.

\subsection{Hypergraph embedding}\label{sec:model:embedding}
The task of embedding the vertices of a  hypergraph into a lower-dimensional \textit{latent space} is a common approach to understanding the relational structure within a hypergraph \citep{zhou_learning_2006, turnbull2019latent}. Here we establish a connection between the Gaussian process latent variable model (\gls{GPLVM}) \citep{lawrence_2003_gplvm} and the Mat\'ern hypergraph \gls{GP} \Cref{equn:GraphGPPrior}, and demonstrate how this synergy can be leveraged to infer a hypergraph latent space embedding in a computationally-efficient manner.

Working under the assumption that our hypergraph can be embedded in a low-dimension \textit{latent space}, the aim here is to learn the position of each vertex in the latent space through a \gls{GP}. We let $X=\{\x_1, \x_2, \ldots, \x_N\},$ represent our $Q$-dimensional latent space, such that the $i^{\text{th}}$ vertex's latent space coordinate is given by $\x_i \in \mathbb{R}^Q$. 

We wish to learn the posterior distribution of the vertices' latent positions conditional on the hypergraph incidence matrix $H$, i.e. $p(X|H) \propto p(H|X)p(X)$. We can use our hypergraph Gaussian process model (Section \ref{sec:model}) to create a mapping between the latent positions and the hypergraph structure using the following model,
\begin{align}
    \label{equn:GPLVMFMarginalised}
    p(H\given X)p(X) = \left( \int p(H\given F)p(F\given X)\mathrm{d}F\right)p(X),
\end{align}
where $F$ is an unobserved matrix with the same dimensions as the incidence matrix $H$, with each column $f_j$ sampled independently from a GP prior, such that 
\begin{align*}
    p(H\given F)  = \cN(F, \sigma^2I_N), \ \ \ p(f_j\given X)  = \cN(\mathbf{0}, \Kxx \KVV) \ \ \mbox{and} \ \ p(X)  = \cN(\mathbf{0}, I_N). 
\end{align*}

The Gram matrix $\Kxx$ is calculated using a Euclidean kernel, e.g., the squared exponential, on the domain $\bbR^Q \times \bbR^Q$, $I_N$ is a $N\times N$ identity matrix, and $\KVV$ is the hypergraph kernel from \Cref{equn:GraphGPPrior}, which uses the hypergraph Laplacian to inform the prior model in an empirical Bayes approach analogous to using a spatial variogram to fix the kernel parameters in a Euclidean model e.g., \cite{diggle1998model}. The incidence matrix $H$ is a binary matrix, however, in this section we make the assumption that $H|F$ follows a Gaussian distribution. This assumption allows us to solve the integral \Cref{equn:GPLVMFMarginalised} analytically,
\begin{align*}
    H|X \sim \mathcal{N}(0,\Kxx \KVV + \sigma^2I_N).
\end{align*}
Without this assumption, approximate inference techniques such as MCMC or variational inference would be required.

The model described here is incredibly flexible as it enables the practitioner to posit their beliefs around the \textit{smoothness} of the latent space through the $\nu$ parameter in \Cref{equn:GraphGPPrior}. Larger values of $\nu$ will result in a more dispersed latent space and larger degrees of separation will be enforced between clusters. Further, covariate information at the vertex level can easily be incorporated into the embedding function by specifying a second kernel that acts across the covariate space of the vertex set and then computing the product of this covariate kernel with the base graph kernel.

\section{Related work}\label{sec:related_work}

To the best of our knowledge, this is the first work to consider \gls{GP} inference for hypergraphs. However, the \gls{SPDE} representation of a Mat\'ern kernel has been successfully used by \cite{lindgren2011explicit} to allow for highly scalable inference on \glspl{GMRF} \citep{rue2005gaussian} and more recently Riemannian manifolds \citep{borovitskiy_2020_riemannian}. Parallel to this line of work, \citep{zhi_2020_gps} derive graph \gls{GP} priors by extending the multi-output \gls{GP} prior given in \cite{vanderWilk_2020_moutput}. Finally, \cite{opolka_2020_graph} derive a deep \gls{GP} prior \citep{damianou_2013_deep} that acts on a vertex set using the spectral convolution of a graph's adjacency matrix, as given by \cite{kipf_2017_semis}. Inference on graphs using GPs, as shown in the aforementioned works, is effective. However, as we show in \Cref{sec:experiments}, such models can be greatly improved by utilising the full hypergraph structure.

Within the deep learning community, \glspl{GNN} \citep{bronstein_2016_geometric} have been studied in the context of hypergraphs from both an algorithmic \citep{feng_2019_hypergraph} and theoretical \citep{bodnar_2021_weisfeiler} perspective. More recently, hypergraph GNN architectures have been extended to incorporate attention modules \citep{zhang_2021_hattention}. Unfortunately, \glspl{GNN} are currently incapable of quantifying the predictive uncertainty on the vertices of a hypergraph. In contrast, the hypergraph \gls{GP} presented here is able to produce well-calibrated uncertainty estimates (see \Cref{sec:exp:peru}).

Analysis of hypergraph data appears much more broadly in the statistics and machine learning literature (see \cite{battiston_networks_2020} for contemporary review), and typical tasks involving hypergraphs include label prediction, classification and clustering \citep{gao2020}. Within the \gls{GP}, the hypergraph informs the dependence structure among the variables associated with the vertices and we note that this departs from a common hypergraph modelling set-up in which there is uncertainty on the hyperedges. Finally, whilst we rely on the hypergraph Laplacian proposed in \cite{zhou_learning_2006}, we note the existence of other Laplacians which may be used in our context (for example, see \cite{chung1993, hu2015, saito2018} for examples), depending on the suitability of the assumptions for a given hypergraph.

\section{Experiments}\label{sec:experiments}
In this section we illustrate the variety of modelling challenges that can be addressed through a hypergraph Gaussian process. A full description of the experimental set up used below can be found in \Cref{app:exp:config}. Further, we release a Python package that enables \gls{GP} inference on hypergraphs using GPFlow\footnote{Licensed under Apache 2.0.} \citep{gpflow_2017_matthews} and HyperNetX \citep{HyperNetX} along with code to replicate the experiments at {\tt https://github.com/RedactedForReview}.

\subsection{Multi-class classification on legislation networks}\label{sec:exp:peru}

In this first section we demonstrate the additional utility that is given from a hypergraph representation by comparing our proposed model to its simpler graph analogue. To achieve this, we consider data that describes the co-sponsorship of legislation within the Congress of the Republic of Peru in 2007 \citep{lee_time-dependent_2017}. To represent this dataset as a hypergraph, we form a hyperedge for each piece of legislation and the constituent vertices correspond to the members of Congress responsible for drafting the respective legislation. In the 2007 Congress there were three distinct groups\footnote{Groups in the Republic of Peru's Congress are collections of political parties}: the government, opposition and minority groups. Our task here is to predict the group affiliation of a member of Congress given only the hypergraph structure. 

\begin{table}[ht]
\centering
    \caption{The performance of the GP models from \Cref{sec:exp:peru} on a graph and hypergraph structure. Results are reported for the 40 held-out vertices, split by the underlying (hyper)graph representation. Bold values denote the best performing model and standard errors are computed across 10 random partitions of the data. For every metric, excluding expected calibration error (ECE), a larger value is better.} 
 \resizebox{\textwidth}{!}{
  \begin{tabular}{lcccc}
  \toprule
    Metric  & Hypergraph  & \multicolumn{1}{p{3cm}}{\centering Graph \\ Binary expansion} &\multicolumn{1}{p{3cm}}{\centering Graph \\ Weighted expansion} \\
    \midrule
    Accuracy & $\mathbf{0.9\pm 0.002}$  &    $ 0.7  \pm 0.015 $ & $ 0.75  \pm 0.009 $ \\ 
    Recall &  $\mathbf{0.89\pm 0.002}$  &   $ 0.74  \pm 0.002 $  &  $ 0.78  \pm 0.006 $ \\ 
    Precision & $\mathbf{0.9\pm 0.002}$ &  $ 0.73  \pm 0.033 $ & $ 0.79  \pm 0.007 $ \\ 
    ECE & $\mathbf{0.13}  \pm \mathbf{0.001} $ &   $0.29\pm 0.003$ & $ 0.2  \pm 0.002 $ \\ 
    Log-posterior density & $\mathbf{-0.34\pm 0.017}$  & $ -0.89  \pm 0.001 $ &$ -0.45  \pm 0.005 $  \\ 
    \bottomrule
    \end{tabular}
    }
    \label{tab:voting_preds}
\end{table}

As far as we are aware, this is the first piece of work to consider \gls{GP} modelling on hypergraphs, so no directly comparative method is available. Therefore, to establish suitable benchmarks, we represent our hypergraph as a graph using both a weighted and binary clique expansion \citep{agarwal_2005_beyond}; a process we briefly outline in \Cref{app:exp:graphRepr}. From thereon, \gls{GP} modelling can be accomplished using the graph kernel provided in \cite{borovitskiy_matern_2020}. Although additional mappings from a hypergraph to a graph are commonly used in the literature, such as the star expansion, we focus on the clique expansion since this preserves the roles of the vertices. 

We employ the categorical likelihood function of \cite{hernandez_2011_robust} and a multi-output \gls{GP} $f: V \rightarrow \mathbb{P}^2$, where $\mathbb{P}^2$ is the probability 2-simplex and each output dimension corresponds to the probability of the respective vertex being attributed to one of the three political groups. As can be seen in \tabref{tab:voting_preds}, the hypergraph representation yields significantly fewer misclassified nodes with equally compelling precision and recall statistics. \glspl{GP} are commonly used due to their ability to quantify predictive uncertainty and, as can be seen by the \gls{ECE} values in \tabref{tab:voting_preds}, the hypergraph representation facilitates substantially improved posterior calibration.

\begin{figure}[ht]
    \centering
    \includegraphics[width=\textwidth]{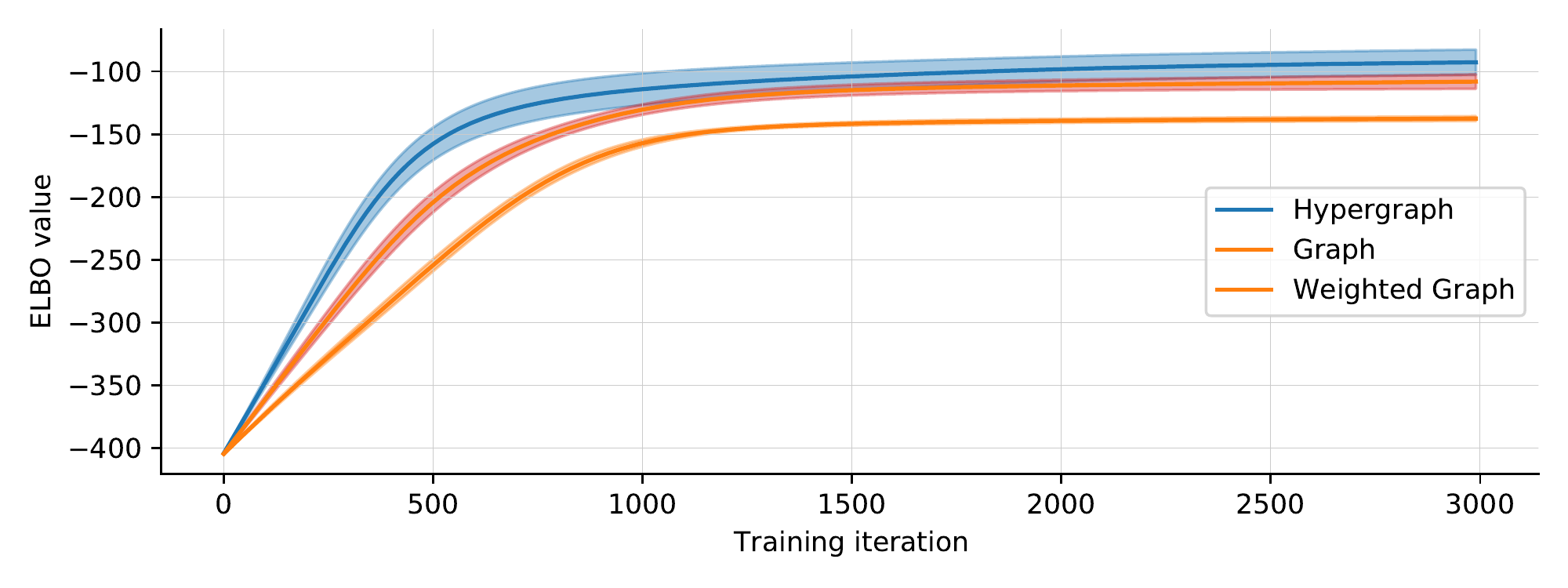}
    \caption{Convergence of the \gls{ELBO} throughout optimisation. A larger value is better. Maximising the \gls{ELBO} is analogous to minimising the Kullback-Leibler divergence from our approximate \gls{GP} to the true \gls{GP}. We can see that representing the data as a hypergraph in our \gls{GP} yields substantially more efficient optimisation, and the resultant model gives a tighter bound on the true log-likelihood.}
    \label{fig:elbo_curves}
\end{figure}

\begin{figure}[h]
    \centering
    \includegraphics[width=\textwidth]{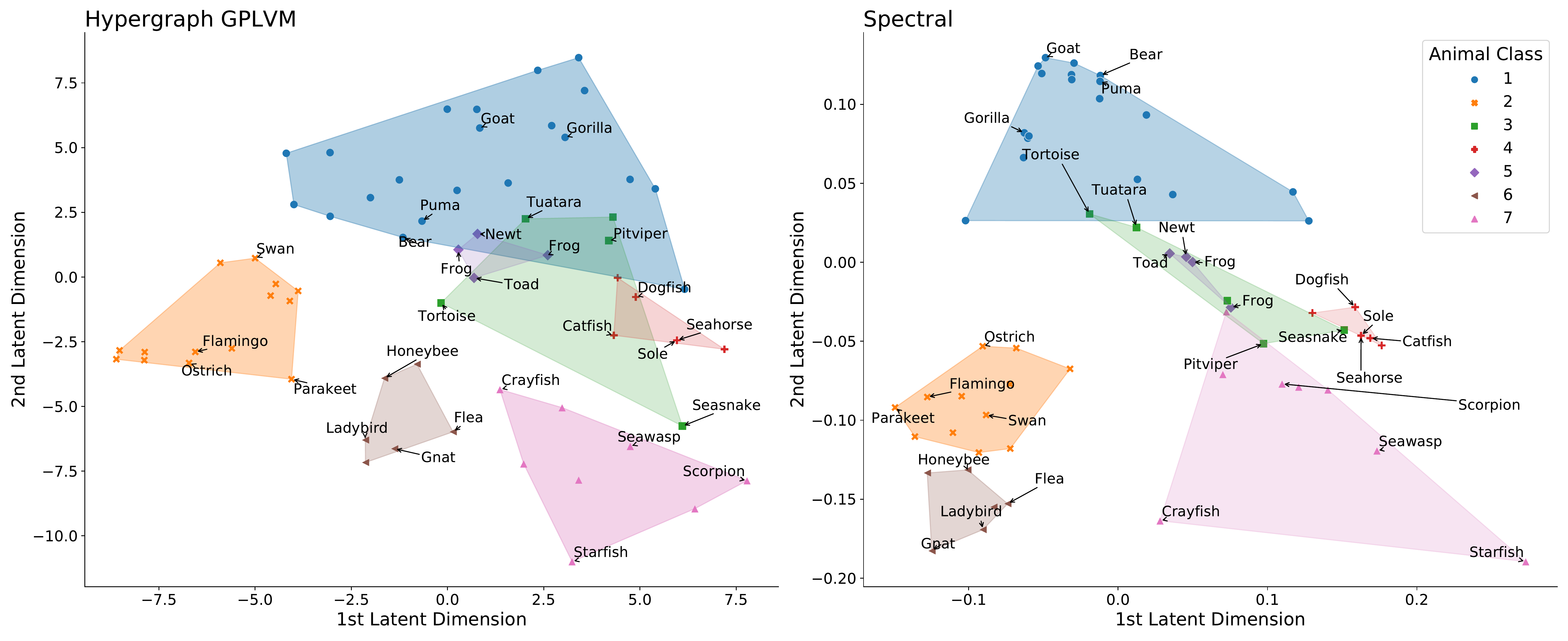}
    \caption{Visualisation of the latent space projections learned using our hypergraph based GPLVM and the method outlined in \cite{zhou_learning_2006} titled \textit{spectral}. Observations are coloured based upon their respective class and the shadings denote the convex hull of the entire class. To avoid a cluttered layout, we randomly label four animals per class and provide fully labelled plot in \Cref{app:results:zoo}.}
    \label{fig:zoo_latent_space}
\end{figure}

In addition to the universally superior predictive measures reported in \tabref{tab:voting_preds}, the \gls{ELBO} curves in \Cref{fig:elbo_curves} show that the hypergraph representation facilitates more efficient optimisation. This result is particularly noteworthy as a common criticism of \glspl{GP} is the computational demands invoked by their optimisation. Therefore, being able to achieve more efficient optimisation is critically important if we hope for these \gls{GP} models to be more widely adopted by machine learning practitioners.

\subsection{Hypergraph latent embeddings}\label{sec:exp:embedding}
In this example we demonstrate the enhanced utility provided by the hypergraph \gls{GPLVM} outlined in \Cref{sec:model:embedding} in comparison to traditional spectral models that exist in the hypergraph literature, namely the work of \cite{zhou_learning_2006}. The dataset under consideration is the Zoo data from the UCI Machine Learning Depository \citep{Dua_2017_UCI} and our goal is to embed the animals' relational structure into a 2-dimensional latent space. We represent the dataset as a hypergraph, such that each vertex corresponds to a specific animal, and each hyperedge corresponds to an animal's attribute e.g., number of legs, presence of a tail. There are seven possible labels within the dataset that roughly describe an animal's taxonomic class. We remove this attribute from the dataset and only use it for the colouring of the vertices' latent space positions in \Cref{fig:zoo_latent_space}. We use a squared exponential kernel to compute the Gram matrix over the latent spaces' coordinates (denoted $\Kxx$ in \Cref{sec:model:embedding}).

We can visually see from \Cref{fig:zoo_latent_space} that the hypergraph \gls{GPLVM} is better able to uncover the latent group structure that is present in the data through the more homogeneous group clustering. Further, using the convex hulls in \Cref{fig:zoo_latent_space} we are able to measure the degree to which each latent space representation has been able to recover the original classes. We report the adjusted mutual information \citep{nguyen_2010_ami}, homogeneity and completeness \citep{rosenberg_2007_vmeasure} in \tabref{tab:clustering_clf} where it can be seen that our hypergraph \gls{GPLVM} outperforms the approach of \cite{zhou_learning_2006} for all considered metrics.

\begin{table}[ht]
    \centering
    \caption{Adjusted mutual information, homogeneity and completeness of the latent spaces produced by our proposed hypergraph \gls{GPLVM} and the spectral embedding of \cite{zhou_learning_2006}. Each score is defined on the domain $[0, 1]$ and a higher score is better. A description of these three measures is provided in \Cref{app:experimental_details}.} 
    \begin{tabular}{lcc}
        \toprule
          & Hypergraph GPLVM & Spectral embedding \\
          \midrule
          Adjusted mutual information & $\mathbf{0.689}$ & $0.606$ \\
          Homoegeneity & $\mathbf{0.774}$ & $0.668$\\
          Completeness & $\mathbf{0.683}$ & $0.633$\\
 \bottomrule
    \end{tabular}
    \label{tab:clustering_clf}
\end{table}

\subsection{Kernelised probabilistic matrix factorisation}\label{sec:exp:kpmf}
In this section we study the effect of a hypergraph kernel when used for \gls{KPMF} \citep{zhou_2012_kpmf}. Given a partially observed matrix $R\in\bbR^{N_U \times N_W}$, for example a matrix of movie ratings, our aim is to predict the missing values of $R$ using the assumption that the outer product of two latent matrices $U\in\bbR^{N_U \times D}$ and $W\in\bbR^{N_W \times D}$ generates $R$ where $D$ is the latent factors' dimension. We place zero-mean \gls{GP} priors over the columns of $U$ and $W$,
\begin{align}
    \label{equn:KPMFPrior}
    U_{:, d} \sim \GP(0, \kappa_U), \qquad W_{:, d} \sim \GP(0, \kappa_W).
\end{align}

A factorised Gaussian likelihood is placed over $R$ such that
\begin{align}
    \label{equn:KMPFLikelihood}
p(R\given U,W, \sigma_n^2) = \prod\limits_{n=1}^N\prod\limits_{m=1}^M \cN\left(R_{n, m}\given U_{n,:}W^{\top}_{m, :}, \sigma^2\right)^{\delta_{n, m}}
\end{align}
where $\delta_{n,m}=1$ if $R_{n,m}$ is not empty, and 0 otherwise. We then learn the optimal pair of latent matrices by optimising the model's log-posterior with respect to $U$ and $W$.

To employ this model we use the MovieLens-100k dataset that consists of 100,000 reviews on 1682 movies from 943 users \citep{harper_2016_movielens}. We define the \gls{GP} priors in \Cref{equn:KPMFPrior} as our hypergraph prior from \Cref{equn:GraphGPPrior}. Letting $R\in\bbR^{943\times1682}$ be the matrix of movie ratings, we can construct the incidence matrix $H$ of the hypergraph for which users represent vertices and the collections of users who review the same movie as hyperedges by calculating $[H]_{n,m}=\delta_{n, m}$ for $1\leq n \leq N_U$ and $1 \leq m \leq N_W$ from \Cref{equn:KMPFLikelihood}. Through the hypergraph's dual representation (see Section \ref{sec:model}), we are able to define incidence matrices where movies are vertices and hyperedges contain the set of movies that an individual user reviewed by $H^{\top}$.

\begin{table}[ht]
    \centering
    \caption{\Gls{RMSE} for our hypergraph \gls{GP} \gls{KPMF} model in \secref{sec:exp:kpmf} and comparative diffusion model from \cite{zhou_2012_kpmf}. We report \gls{RMSE} on the held-out data where a lower \gls{RMSE} indicates a better model. All models are fit to datasets whereby 20\% and 80\% of the data is held back for testing. The sparse variant of our model uses an inducing point set with size equal to $\nicefrac{1}{20}$th of the $U$ and $W$ matrices. Standard errors reported here are across 10 random partitions of the data and are significant after the third decimal place.} 

    \begin{tabular}{lccc}
        \toprule
        Testing proportion & Diffusion & 
        \thead{Hypergraph KPMF \\ Full } &
        \thead{Hypergraph KPMF \\ Sparse} \\
          \midrule
          20\% & $1.39 \pm 0.0$ & $\mathbf{1.21 \pm 0.0}$ & $1.40 \pm 0.1$  \\
          80\% & $1.51\pm0.0$ & $\mathbf{1.27 \pm 0.0}$ & $1.56\pm 0.1$\\
 \bottomrule
    \end{tabular}
    \label{tab:kpmf_results}
\end{table}

In \tabref{tab:kpmf_results} we compare our hypergraph-based \gls{KPMF} to the \gls{KPMF} model given in \cite{zhou_2012_kpmf} which uses a diffusion kernel \citep{kondor_2002_diffusion} that we describe fully in \Cref{app:details:kpmf}. Further, we test the effect of our inducing point scheme described in \Cref{sec:model:inducingPoints} by approximating $\kappa_U$ and $\kappa_W$ with 45 and and 80 inducing vertices respectively. As can be seen by the smaller \gls{RMSE} in \tabref{tab:kpmf_results}, the hypergraph kernel offers substantial improvements in comparison to the regular \gls{KPMF} model. 
Whilst attaining a larger \gls{RMSE} than the dense hypergraph \gls{KPMF} model, the sparse approximation is able to compute \Cref{equn:KMPFLikelihood} 4.3 times faster than the both diffusion and dense \gls{KPMF} models. We note that a sparse approximation is not strictly necessary in this example, however, selecting a dataset where fitting full rank kernels is possible allows us to quantitatively assess the difference in predictive RMSEs. Furthermore, this sparse approximation allows us to scale this model to datasets where $U$ and/or $W$ contains millions of observations. In such a setting, fitting a dense kernel would yield an intractable likelihood \Cref{equn:KMPFLikelihood} due to the requirement of inverting $\kappa_U$ and $\kappa_W$. However, using our sparse kernel approximation would permit scaling to datasets of this size.

\section{Concluding remarks}\label{sec:conclusions}
\paragraph{Limitations}Whilst we have found the hypergraph Laplacian from \Cref{equn:HypergraphLaplacian} to perform well experimentally, we are aware of several criticisms of this in the literature \citep{agarwal_2005_beyond, ren_2008_spectral}. Despite the limitations this could cause, we see this as an illuminating potential pathway for future research through the consideration of alternative hypergraph Laplacians.

Finally, we only consider hypergraphs that are fixed in the number of nodes and hyperedges. Whilst valid for the use cases we consider in \Cref{sec:experiments}, there exists a large number of hypergraph applications and methodologies whereby the number of vertices and/or hyperedges is either evolving or considered unknown.

\paragraph{Final remarks}In this work we consider the task of performing inference on the vertices of a general hypergraph. To do so, we establish a connection with the Mat\'ern graph kernel and provide a detailed framework for positing the rich structure of a hypergraph into this kernel. We further provide a framework for embedding high-dimensional hypergaphs into a lower-dimensional latent space through GPLVMs as well as a principled approach to selecting inducing vertices for sparse hypergraph GPs. Finally, we demonstrate the enhanced utility of our work through three real-world examples, each of which illustrate distinct aspects of our framework. We envision that the techniques presented in this work are of high utility to both the \gls{GP} and graph theory communities and will facilitate the blossoming of future research at the intersection of these two areas.

\clearpage

\bibliographystyle{custom}
\bibliography{main}

\clearpage

\begin{appendix}
\section{Full experimental details}\label{app:experimental_details}
\subsection{Training configurations}\label{app:exp:config}
\paragraph{Graph representation}In \Cref{sec:exp:peru} and \ref{sec:exp:kpmf} our hypergraph kernel is compared against kernels operating on regular graphs. To represent our hypergraph as a graph we use a clique expansion (\Cref{app:exp:graphRepr}). In \Cref{sec:exp:peru} we use both a weighted and binary clique expansion and, due to its superior performance, only a weighted expansion in \Cref{sec:exp:kpmf} to compute the diffusion kernel. 

\paragraph{Hardware}All experiments are carried out on an Nvidia Quadro GP100 GPU, a 20 core Intel Xeon 2.30GHz CPU and 32GB of RAM.

\paragraph{Implementation}All code is implemented using GPFlow \citep{gpflow_2017_matthews} and is made publicly available at \\ {\tt https://github.com/RedactedForReview}.

\paragraph{Optimisation}For the experiments carried out in \Cref{sec:exp:peru} and \Cref{sec:exp:embedding} we use natural gradients \citep{amari_98_natural} for optimisation of the variational parameters with a learning rate of 0.001 as per \cite{salimbeni_2018_natural}. For optimisation of the \gls{GP} models' hyperparameters in all experiments, we use the Adam optimiser with the recommended learning rate of 0.001 \citep{kingma_2015_adam}.

\paragraph{Parameter constraints} For all parameters where positivity is a constraint (i.e. variance), the softplus transformation i.e., $\log(1+\exp(x))$. Inference is then conducted on the unconstrained parameter, however, we report the re-transformed parameter i.e. the constrained representation.

\paragraph{Parameter initialisation}For the graph kernels we initialise the smoothness, lengthscale and variance parameters to 1.5, 5.0 and 1.0 respectively. For the Gaussian likelihood used in \Cref{sec:exp:embedding}, the variance is initialised to 0.01.

\paragraph{Reported metrics}Given the true test observation $y$ and the predicted value $\hat{y}$, we recall the following definitions: \gls{FN}: $\mathds{1}_{\hat{y}=1\given y = 0}$, \gls{TN}: $\mathds{1}_{\hat{y}=0\given y = 0}$, \gls{FP}: $\mathds{1}_{\hat{y}=0\given y = 1}$, and \gls{TP}: $\mathds{1}_{\hat{y}=1\given y = 1}$. Using these definitions, we report the following metrics across the experiments in \Cref{sec:exp:peru} for the $N$ heldout observations:
\begin{align}
    \text{Accuracy:} \quad & \TruePos + \FalsePos \nonumber \\
    \text{Precision}: \quad & \frac{\TruePos}{\TruePos + \FalsePos} \nonumber \\
    \text{Recall}: \quad & \frac{\TruePos}{\TruePos + \FalseNeg}. \nonumber 
\end{align}
In addition to these, we also report the \gls{ECE} \citep{naeini_2015_bayesian}
\begin{align}
    \text{ECE:} \quad & \mathbb{E}\left[\operatorname{Pr}\left(\hat{y}=y\given \hat{p} = p\right) -p \right] \nonumber 
\end{align}
where $\hat{p}$ is predicted mean.

For the latent space embedding carried out in \Cref{sec:exp:embedding} we use three metrics to quantify the quality of the learned latent space: adjusted mutual information, homogeneity and completeness. \textit{Adjusted mutual information} \citep{nguyen_2010_ami} is an information theoretic measure that quantifies the degree to which the latent class labels and those generated by the convex hull of each class' latent space coordinates align. Additionally, \textit{homogeneity} measures how homogeneous each cluster's constituent vertices are, whilst \textit{completeness} measures the degree to which all class members were assigned to the correct cluster \citep{rosenberg_2007_vmeasure}.

\FloatBarrier

\subsection{Clique expansion}\label{app:exp:graphRepr}
Considering a hypergraph $\cG$, we can build a regular graph representation using a clique expansion. Let $H$ be the hypergraph's incidence matrix and $A_w$ and $A_b$ be the graph's adjacency matrix for the weighted and binary clique expansion used in \Cref{sec:exp:peru}, respectively. The  $(i, j)^{\text{th}}$ entry of $A_w$ and $A_b$ can be written as
\begin{alignat}{2}
    \text{Weighted:} \quad & [A_w]_{ij} = && \sum_{e\in\cG}\mathbbm{1}(i, j \in \cG) \label{equn:w_clique} \\
    \text{Binary:}  \quad & [A_b]_{ij} = && \sum_{e \in \cG}\mathbbm{1}(w_{ij}>0) \label{equn:b_clique}
\end{alignat}

\paragraph{Example:}Consider the following incidence matrix 
\begin{align}
\label{equn:exampleInc}
    H = \left[\begin{array}{cccc}
    1&0&1&0\\
    1&1&1&1\\
    1&0&1&1\\
    1&1&1&1\\
    0&1&1&0\\
    \end{array}\right]
\end{align}
Using the clique expansions in \Cref{equn:w_clique} and \Cref{equn:b_clique}, the corresponding adjacency matrix $A_w$ and $A_b$ can be written as 
\begin{align}
\label{equn:exampleAdj}
\quad A_{w}=\left[\begin{array}{ccccc}
    0&2&2&2&1\\
    2&0&3&4&2\\
    2&3&0&3&1\\
    2&4&3&0&2\\
    1&2&1&2&0\\
    \end{array}\right], \quad A_{b}=\left[\begin{array}{ccccc}
    0&1&1&1&1\\
    1&0&1&1&1\\
    1&1&0&1&1\\
    1&1&1&0&1\\
    1&1&1&1&0\\
    \end{array}\right]
\end{align}

\begin{figure}[ht]
    \centering
    \includegraphics[width=\textwidth]{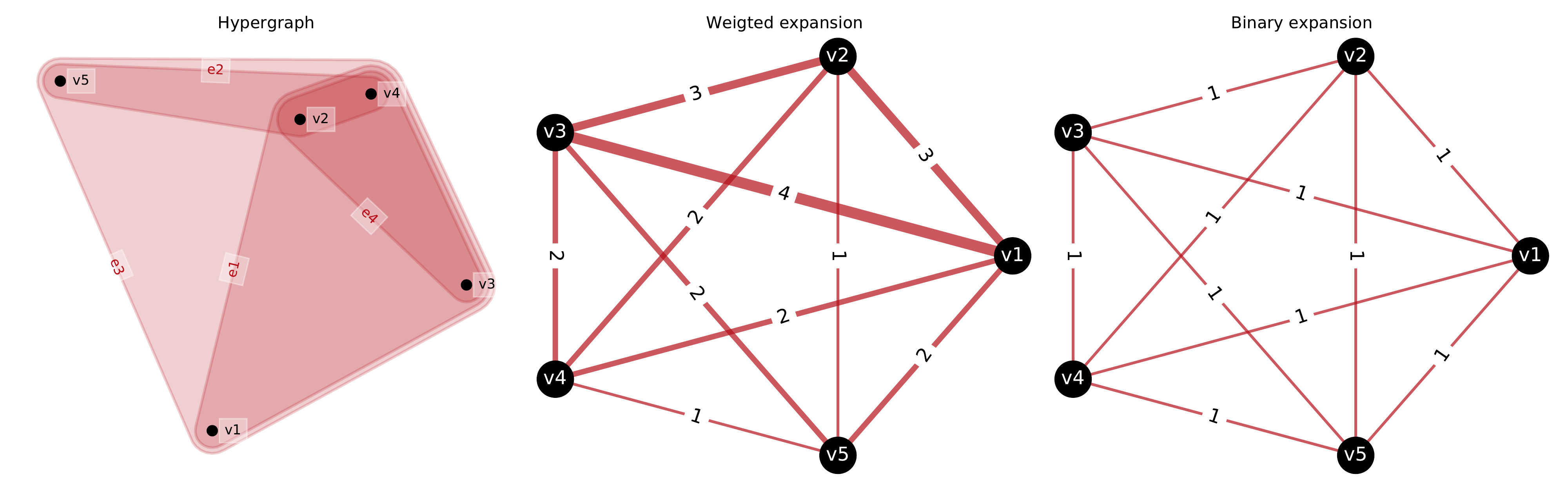}
    \caption{A depiction of the incidence matrix $H$ in \Cref{equn:exampleInc} and clique expanded adjacency matrices $A_w$ and $A_b$ stated from \Cref{equn:exampleAdj}. The width of the edges in the graph are proportional to the edge's weight.}
    \label{fig:hypergraph_to_graph}
\end{figure}

A depiction of the matrix representations in \Cref{equn:exampleInc} and \Cref{equn:exampleAdj} is given in \Cref{fig:hypergraph_to_graph}.

\subsection{Diffusion kernel}\label{app:details:kpmf}
The corresponding Gram matrix $K_D$ of the diffusion kernel \citep{kondor_2002_diffusion} can be written as 
\begin{align}
    \label{equn:DiffusionKernel}
    K_D = \lim_{n\rightarrow\infty}\left(1-\frac{\beta \Delta}{n}\right)^n
\end{align}
where $\beta\in\bbR$ is the bandwidth parameter and $\Delta$ is the (hyper)graph Laplacian matrix. Intuitively, the $(i,j)^{\text{th}}$ entry of $K$ can be seen as the amount of substance that has diffused from $v_i$ to $v_j$ in the original graph. Letting $\beta=0$ corresponds to zero diffusion and larger values of $\beta$ will give larger amounts of diffusion. As advised in \cite{zhou_2012_kpmf}, we set $\beta=0.01$ in \Cref{sec:exp:kpmf}.

Computation of the diffusion kernel given in \Cref{equn:DiffusionKernel} can be achieved through
\begin{align}
    \label{equn:DiffusionKernelComp}
    K_D = \exp(-\beta \Delta),
\end{align}
as given in \cite{kondor_2002_diffusion}.

\clearpage

\section{Additional experimental results}\label{app:results}

\subsection{Latent space embedding}\label{app:results:zoo}
\rotatebox{90}{\begin{minipage}{0.85\textheight}
    \includegraphics[width=\textwidth]{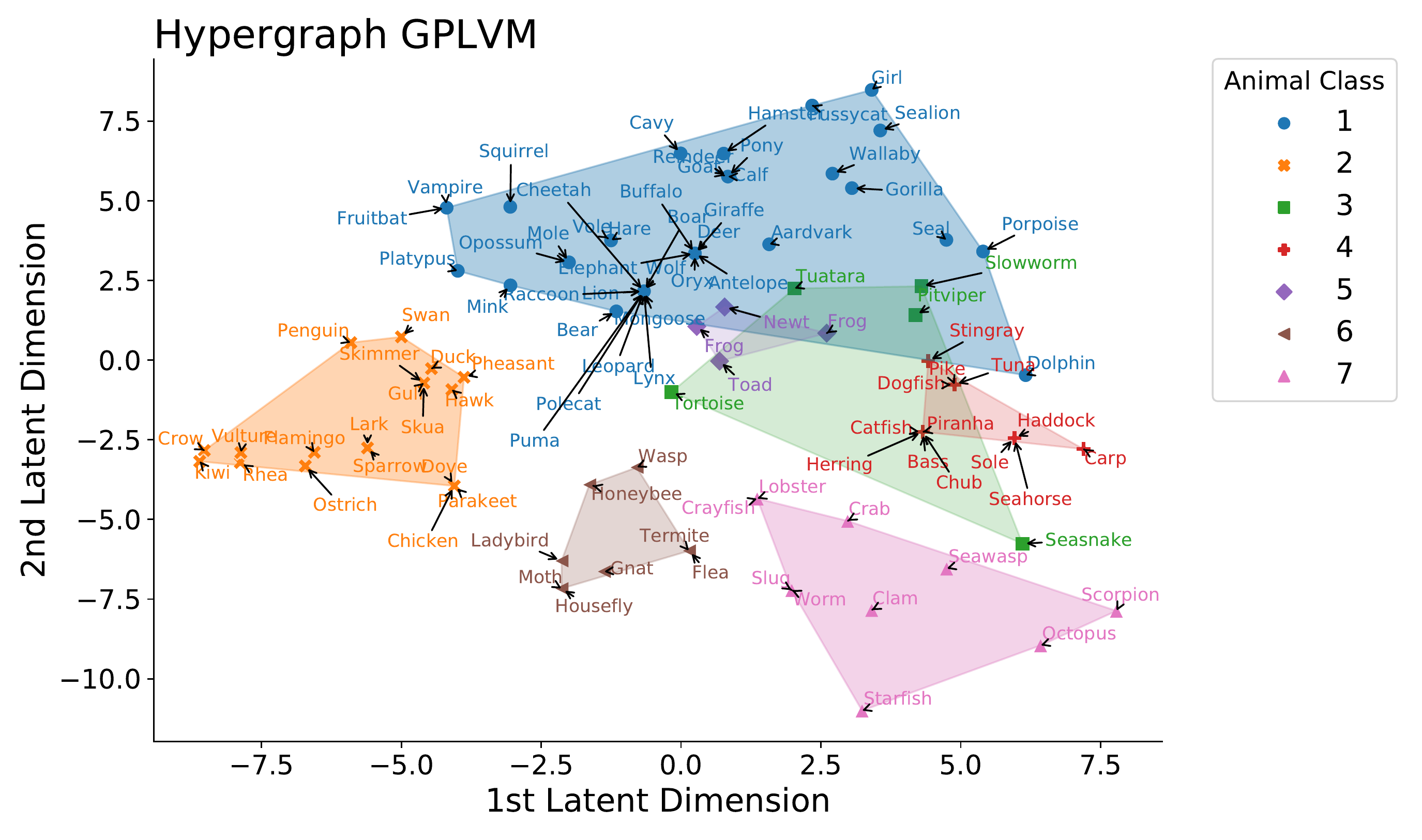}
    \captionof{figure}{A replication of the hypergraph GPLVM plot given in \Cref{fig:zoo_latent_space} with every vertex labelled.}
    \label{fig:gplvm_latent_space_full}
\end{minipage}}

\clearpage

\rotatebox{90}{\begin{minipage}{0.9\textheight}
    \includegraphics[width=\textwidth]{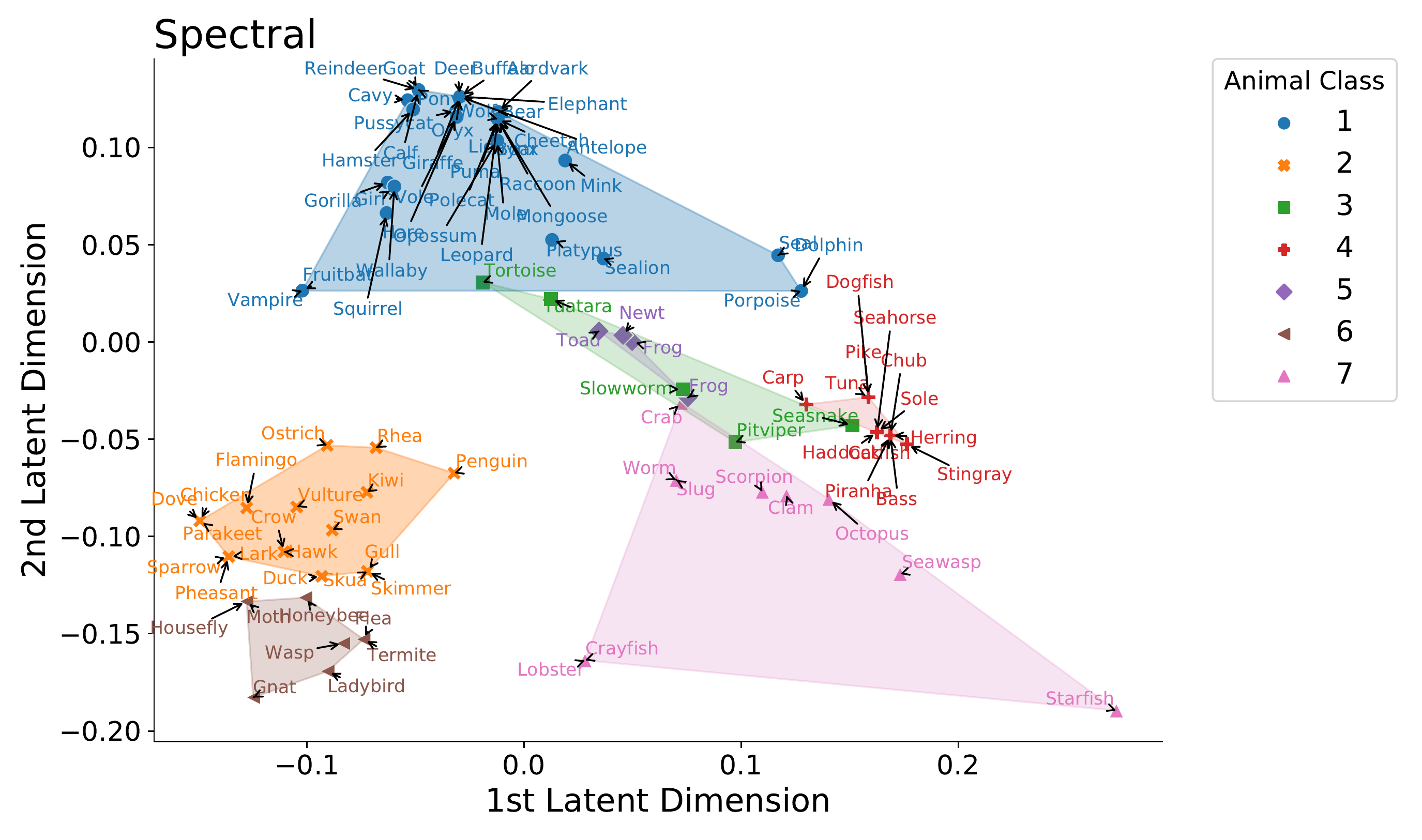}
    \captionof{figure}{A replication of the spectral plot given in \Cref{fig:zoo_latent_space} with every vertex labelled.}
    \label{fig:spectral_latent_space_full}
\end{minipage}}

\end{appendix}

\end{document}